\documentclass[10pt,twocolumn,letterpaper]{article}

\usepackage{cvpr}
\usepackage{times}
\usepackage{epsfig}
\usepackage{graphicx}
\usepackage{amsmath}
\usepackage{amssymb}

\usepackage{bm} 
\usepackage{multirow} 
\usepackage{hhline} 
\usepackage{array} 
\newcolumntype{M}[1]{>{\centering\arraybackslash}m{#1}} 
\usepackage[utf8]{inputenc}
\usepackage[english]{babel}
\usepackage[dvipsnames]{xcolor}
\definecolor{myred}{RGB}{242,150,153}
\definecolor{mygreen}{RGB}{153,206,152}

\usepackage[pagebackref=true,breaklinks=true,letterpaper=true,colorlinks,bookmarks=false]{hyperref}

\cvprfinalcopy 

\def\cvprPaperID{940} 

\begin{document}

\title{Hi-CMD: Hierarchical Cross-Modality Disentanglement for Visible-Infrared Person Re-Identification}

\author{Seokeon Choi  \quad Sumin Lee \quad Youngeun Kim \quad Taekyung Kim \quad Changick Kim\\
Korea Advanced Institute of Science and Technology, Daejeon, Korea\\
{\tt\small $\{$seokeon, suminlee94, youngeunkim, tkkim93, changick$\}$@kaist.ac.kr}
}
\maketitle

\begin{abstract}

Visible-infrared person re-identification (VI-ReID) is an important task in night-time surveillance applications, since visible cameras are difficult to capture valid appearance information under poor illumination conditions. Compared to traditional person re-identification that handles only the intra-modality discrepancy, VI-ReID suffers from additional cross-modality discrepancy caused by different types of imaging systems. To reduce both intra- and cross-modality discrepancies, we propose a Hierarchical Cross-Modality Disentanglement (Hi-CMD) method, which automatically disentangles ID-discriminative factors and ID-excluded factors from visible-thermal images. We only use ID-discriminative factors for robust cross-modality matching without ID-excluded factors such as pose or illumination. To implement our approach, we introduce an ID-preserving person image generation network and a hierarchical feature learning module. Our generation network learns the disentangled representation by generating a new cross-modality image with different poses and illuminations while preserving a person's identity. At the same time, the feature learning module enables our model to explicitly extract the common ID-discriminative characteristic between visible-infrared images. Extensive experimental results demonstrate that our method outperforms the state-of-the-art methods on two VI-ReID datasets. The source code is available at: \href{https://github.com/bismex/HiCMD}{https://github.com/bismex/HiCMD}.

\end{abstract}

\section{Introduction}

Person re-identification (ReID) aims to match a specific person across multiple non-overlapping camera views. Due to its usefulness in security and surveillance systems, person ReID has been of great research interest in recent years. Existing ReID methods mainly treat visible images captured by single-modality cameras, and depend on human appearance for RGB-RGB matching \cite{reid_cvpr2019_PAUL, reid_cvpr2019_CASN, reid_cvpr2019_PATB, reid_cvpr2019_DIMN, reid_cvpr2019_RAM, reid_cvpr2019_SPT}. However, visible light cameras can not capture all the appearance characteristics of a person under poor illumination conditions. For these conditions, most surveillance cameras automatically switch from visible to the infrared mode in dark environments \cite{vireid_iccv2017_SYSU, sensor}. After all, it becomes essential to consider visible-infrared person re-identification (VI-ReID). The goal of VI-ReID is to match pedestrians observed from visible and infrared cameras with different spectra.

\begin{figure}[t]
\begin{center}
\includegraphics[width=1.0\linewidth]{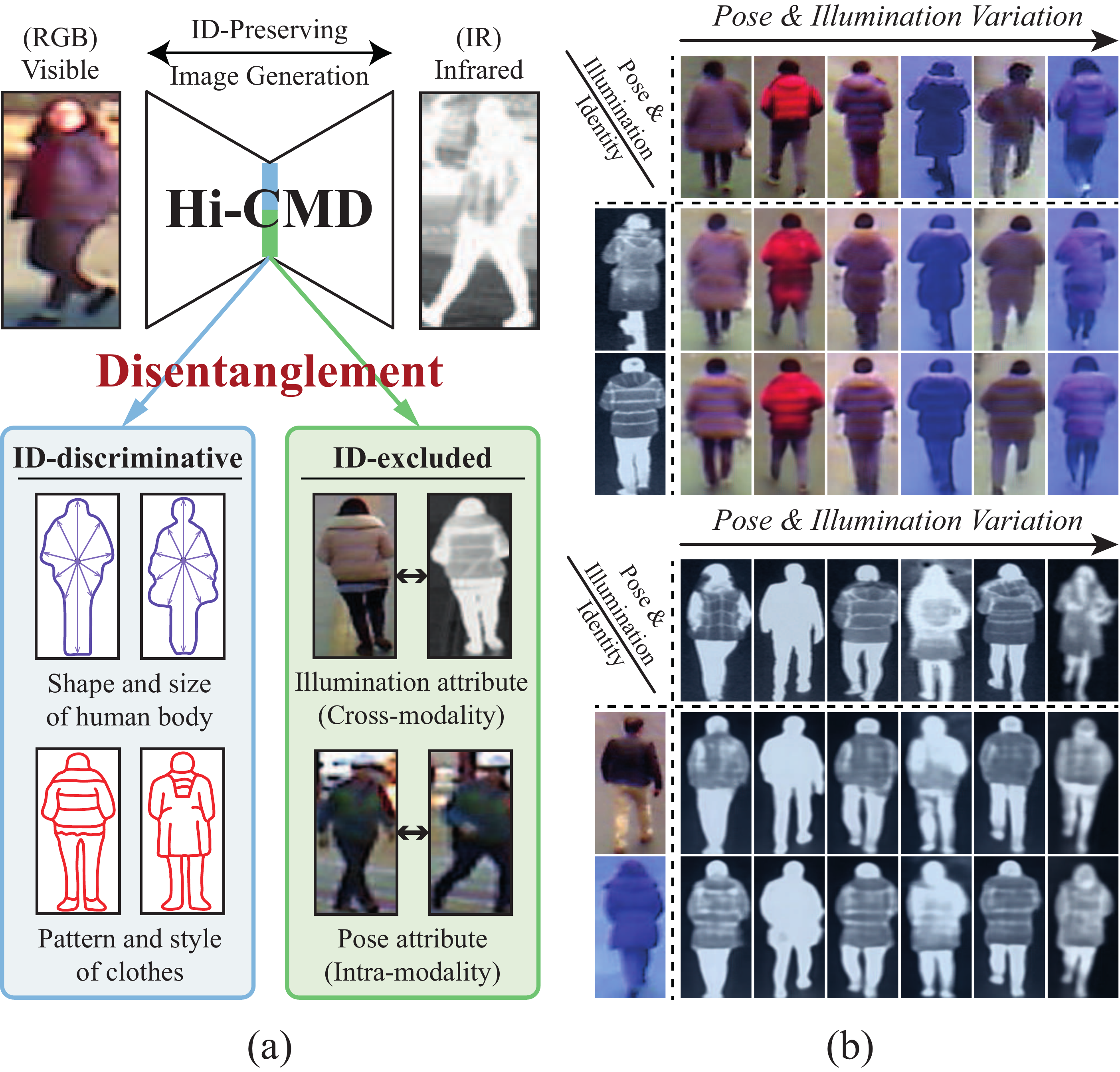}
\end{center}
\vspace{-2.2mm}
\caption{(a) Illustration of our Hierarchical Cross-Modality Disentanglement (Hi-CMD) concept. Our Hi-CMD method aims to hierarchically disentangle ID-discriminative factors (\eg body shape and clothes pattern) and ID-excluded factors (\eg pose and illumination) from RGB-IR images. (b) Examples of ID-preserving Person Image Generation (ID-PIG). The images in each row show that pose and illumination attributes can be changed while maintaining the identity information. Best viewed in color.}
\vspace{-1.2mm}
\label{fig:concept}
\end{figure}
%



Compared to the traditional ReID task that only has the intra-modality discrepancy problem, VI-ReID encounters the additional cross-modality discrepancy problem resulting from the natural difference between the reflectivity of the visible spectrum and the emissivity of the thermal spectrum \cite{rgb-ir}. Eventually, the coexistence of intra- and cross-modality discrepancies leads to a critical situation where the intra-class distance is larger than the inter-class distance in VI-ReID \cite{vireid_tifs2019_eBDTR, vireid_cvpr2019_D2RL}. In this situation, most studies \cite{vireid_iccv2017_SYSU, vireid_aaai2018_TONE, vireid_ijcai2018_BCTR, vireid_ijcai2018_cmGAN, vireid_tifs2019_eBDTR, vireid_aaai2019_HSME} have attempted to reduce both discrepancies with feature-level constraints like the traditional ReID methods \cite{PCB, SVDNET}. It is difficult to eliminate the intractable discrepancies successfully using only feature-level constraints, since the illumination and pose attributes are entangled in a single image. More recent research \cite{vireid_cvpr2019_D2RL} has attempted to bridge the cross-modality gap using an image-level constraint. However, they only translate an infrared (or visible) image into its visible (or infrared) counterpart without considering intra-modality discrepancy, despite the insufficient amount of cross-view paired training data. 

To mitigate the coexisting intra- and cross-modality discrepancies at the same time, we propose a novel Hierarchical Cross-Modality Disentanglement (Hi-CMD) method, as shown in Fig. \ref{fig:concept} (a). The goal of our approach is to hierarchically disentangle ID-excluded factors (\ie pose and illumination) and ID-discriminative factors (\ie body shape and clothes pattern) from cross-modality images using image-level constraints. To this end, we introduce the ID-preserving Person Image Generation (ID-PIG) network. The ID-PIG network focuses on learning the ID-excluded feature representation by replacing some latent vectors in a pair of cross-modality images. As a result, ID-PIG can transform pose and illumination attributes while preserving the identity information of a person as visualized in Fig. \ref{fig:concept} (b). The visualization results of the ID-PIG network show that unnecessary information (\ie pose or illumination attribute) can be separated from entangled representations.




Besides, we introduce the Hierarchical Feature Learning (HFL) module coupled with the ID-PIG network. This module enables the encoders in the generator to extract the common ID-discriminative factors explicitly, which is robust to pose and illumination variations. It also implicitly helps our model separate pose and illumination from RGB-IR images, which improves image generation quality. Finally, the ID-discriminative feature is used for solving the cross-modality image retrieval problem in VI-ReID. Note that we train a whole network in an end-to-end manner without pose-related supervision (\eg 3D skeletal pose, keypoint heatmaps, and pose-guided parsing) compared to the existing pose generation methods \cite{I2Ireid_cvpr2018_trans, I2Ireid_eccv2018_PNGAN, I2Ireid_nips2018_FDGAN, reid_cvpr2019_PATB, reid_cvpr2019_SPT}.

Our main contributions can be summarized as follows:
\begin{itemize}
\item[$\bullet$] We propose a Hierarchical Cross-Modality Disentanglement (Hi-CMD) method. It is an efficient learning structure that extracts pose- and illumination-invariant features for cross-modality matching. To the best of our knowledge, this is the first work to disentangle ID-discriminative factors and ID-excluded factors simultaneously from cross-modality images in VI-ReID.
\item[$\bullet$] The proposed ID-preserving Person Image Generation (ID-PIG) network makes it possible to change the pose and illumination attributes while maintaining the identity characteristic of a specific person. Exploring person attributes through ID-PIG demonstrates the effectiveness of our disentanglement approach. 
\item[$\bullet$] Extensive experimental results show that our novel framework outperforms the state-of-the-art methods on two VI-ReID datasets. The visualization results of the ID-PIG network demonstrate the overwhelming performance of our proposed method.

\end{itemize}

\section{Related Work}

\textbf{Visible-infrared person re-identification.} The visible-infrared person re-identification (VI-ReID) is about matching cross-modality images under different illumination conditions. The VI-ReID task is challenging due to cross-modality variation in addition to intra-modality variation. At the beginning of the study, most of the work has focused on how to design a feature embedding network such as a deep zero-padding network \cite{vireid_iccv2017_SYSU} and a two-stream CNN network \cite{vireid_aaai2018_TONE}. Recently, adversarial learning \cite{vireid_ijcai2018_cmGAN} or metric learning methods \cite{vireid_ijcai2018_BCTR, vireid_aaai2019_HSME, vireid_tifs2019_eBDTR} are applied to learn the feature representation of heterogeneous person images involving intra-modality and cross-modality variations. However, it is difficult to overcome pixel-level differences resulting from illumination or pose variations by feature-level constraints alone because of the insufficient data. In contrast to most existing feature-level approaches, our Hi-CMD method focuses on an image-level approach by combining an image generation technique with the VI-ReID task to effectively bridge both cross-modality and intra-modality gaps.


\begin{figure*}[t]
\begin{center}
\includegraphics[width=1.0\linewidth]{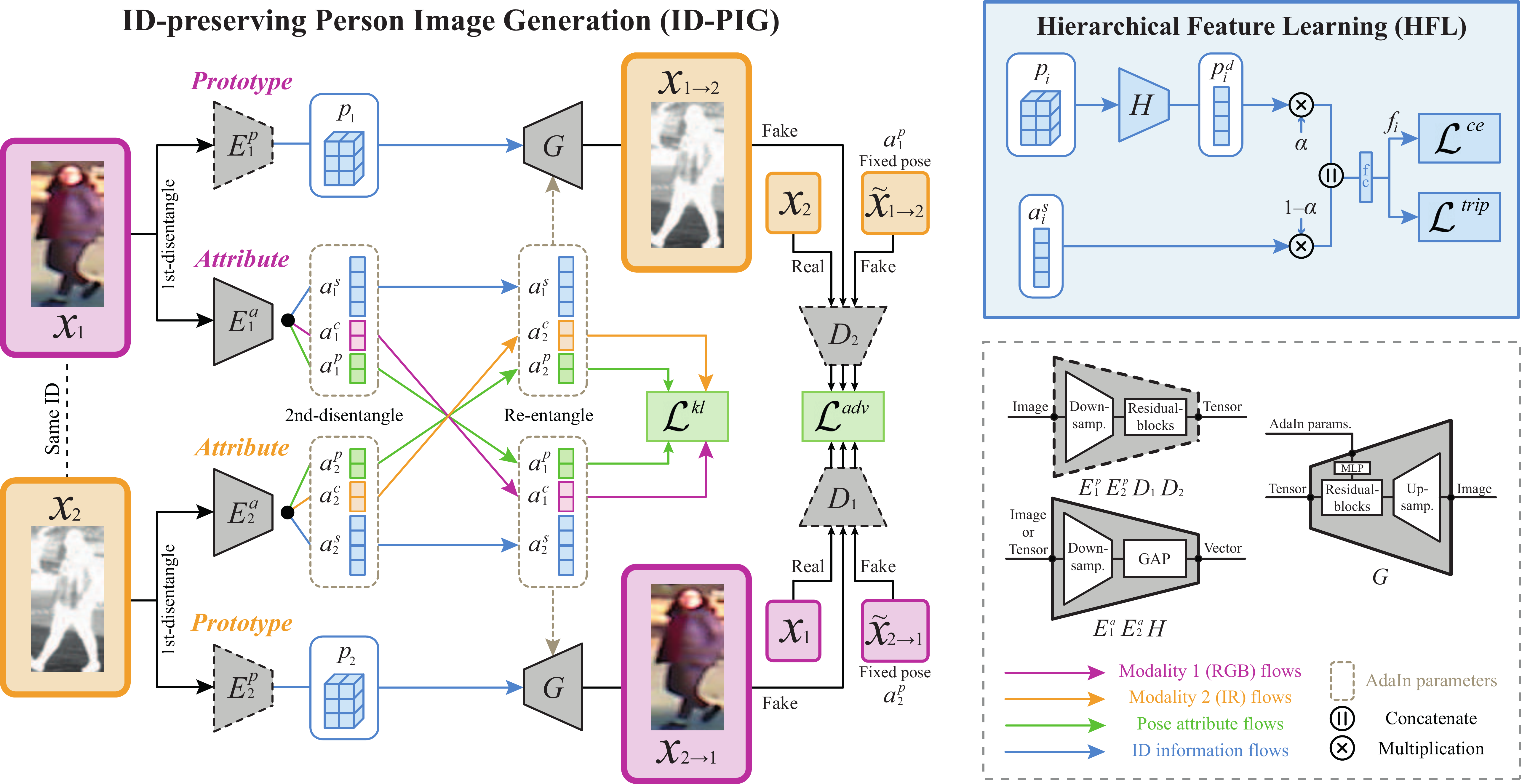}
\end{center}
\vspace{-0.3mm}
\caption{The framework of our Hi-CMD method. The entire framework includes two important components: the ID-preserving Person Image Generation  (ID-PIG) network and the Hierarchical Feature Learning (HFL) module. Our goal is to disentangle ID-discriminative factors and ID-excluded factors from cross-modality images. Reconstruction losses are shown in Fig. \ref{fig:loss}. Best viewed in color.}
\vspace{-0.3mm}
\label{fig:framework}
\end{figure*}

\textbf{Person re-identification based on image generation.} Recently, image generation methods by Generative Adversarial Networks (GANs) \cite{GAN} have drawn a lot of attention in person ReID. Most of the existing work is categorized in two ways as follows: pose transfer \cite{I2Ireid_cvpr2018_trans, I2Ireid_eccv2018_PNGAN, I2Ireid_nips2018_FDGAN} and style transfer \cite{I2Ireid_cvpr2018_PTGAN, I2Ireid_cvpr2019_ATNET, I2Ireid_cvpr2018_SPGAN, vireid_cvpr2019_D2RL}. The work of the former approach points out that the existing datasets do not provide sufficient pose coverage to learn a pose-invariant representation. Accordingly, it addresses this issue by using pose-rich data augmentation. However, since this work is designed for a single-modality environment, it is difficult to apply the pose-guided methods directly to the VI-ReID task.

Another image generation approach in ReID is to reduce the domain gap between different camera domains based on unsupervised domain adaptation \cite{I2Ireid_cvpr2018_PTGAN, I2Ireid_cvpr2019_ATNET, I2Ireid_cvpr2018_SPGAN}. Most methods focus on transforming styles while maintaining the structural information of the person. In a similar approach, Wang \etal \cite{vireid_cvpr2019_D2RL} translate an infrared (or visible) image into its visible (or infrared) counterpart to reduce the cross-modality discrepancy in the VI-ReID task. However, since most style transfer methods in ReID do not care about the lack of cross-view paired training data, view-invariant representations are hardly exploited.


In short, the above image generation methods in ReID handle only structural information or only convert the image style at the image-level. Unlike single-modality person re-identification, it is more important to consider intra-modality and cross-modality characteristics simultaneously in VI-ReID. To this end, we attempt to alleviate the cross-modality and intra-modality discrepancies at the same time by applying a novel hierarchical disentanglement approach even without pose supervision.




\textbf{Disentangled representation learning for recognition.} 
The goal of disentangled representation learning is to extract explanatory factors from diverse data variation for generating a meaningful representation. Recently, considerable attention has been focused on learning disentangled representations in various fields \cite{disentangle_eccv2018_MUNIT, disentangle_nips2018_CDD, disentangle_cvpr2019_styleGAN, disentangle_cvpr2019_fineGAN}. Also in the recognition task, several studies have tried to disentangle the identity-related information and the identity-irrelevant information from an image (\eg pose, viewpoint, age, and other attributes) \cite{disentangle_cvpr2019_gait, disentangle_cvpr2017_DRGAN, disentangle_cvpr2018_preserving, disentangle_cvpr2019_face_age}.  Among them, some previous works in the single-modality person re-identification task have been conducted with a purpose of disentangling foreground, background, and pose factors \cite{disentangle_cvpr2018_reid}, or extracting illumination-invariant features \cite{disentangle_arxiv2019_illumination}. Note that the VI-ReID task is particularly challenging to disentangle the common identity information and the remaining attributes from RGB-ID images due to the coexistence of cross-modality and intra-modality discrepancies. To deal with both pose and illumination attributes simultaneously, we introduce a novel hierarchical disentanglement approach. To the best of our knowledge, this is the first work to disentangle ID-discriminative factors and ID-excluding factors (\ie pose and illumination attributes) from RGB-IR images in the VI-ReID task.

%

%


\vspace{0.5mm}
\section{Proposed Method}
\vspace{0.15mm}
\subsection{Problem Definition and Overview}
\vspace{0.15mm}

\textbf{Problem definition.} We denote the visible image and the infrared image as $\bm{x}_1 \in \mathbb{R} ^ {H \times W \times 3}$ and $\bm{x}_2 \in \mathbb{R} ^ {H \times W \times 3}$ respectively, where $H$ and $W$ are the height and the width of images. Each of images $\bm{x}_1$ and $\bm{x}_2$ corresponds to an identity label $y \in \left\{ 1, 2, ... , N \right\}$, where $N$ is the number of person identities. In the training stage, a feature extraction network $\phi(\cdot)$ is trained with the cross-modality image sets $\mathcal{X}_1$ and $\mathcal{X}_2$. In the testing stage, given a query image with one modality, a ranking list within the gallery set of the other modality is calculated. The distance between two feature vectors $\phi(\bm{x}_1)$ and $\phi(\bm{x}_2)$ is computed by the Euclidean distance.


\textbf{Framework overview.} In the VI-ReID task, the most challenging issue is that both cross- and intra-modality discrepancies coexist between visible and infrared images. To address this issue effectively, we propose a novel Hierarchical Cross-Modality Disentanglement (Hi-CMD) method. Our Hi-CMD method aims to disentangle ID-discriminative factors and ID-excluded factors from cross-modality images to reduce the cross-modality and intra-modality discrepancies at the same time. To achieve this goal, we introduce two key components, the ID-preserving Person Image Generation (ID-PIG) network and the Hierarchical Feature Learning (HFL) module, as shown in Fig. \ref{fig:framework}.



\subsection{Identity Preserving Person Image Generation} \label{sec:ID-PIG}

\textbf{Hierarchical representation.} We present the hierarchical representation of person images for VI-ReID. As illustrated in Fig. \ref{fig:framework}, our ID-PIG network consists of two disentanglement stages. In the first stage, we design a prototype encoder $E_i^{p}$ and an attribute encoder $E_i^{a}$ for each modality ($i\!=\!1$ for visible images and $i\!=\!2$ for infrared images). These encoders $E_i^{p}$ and $E_i^{a}$ map $\bm{x}_i$ to the corresponding prototype code $\bm{p}_i$ and attribute code $\bm{a}_i$, respectively. The prototype code $\bm{p}_i$ is a tensor containing the underlying form of a person appearance such as clothes pattern and body shape. On the other hand, the attribute code $\bm{a}_i$ is a vector including clothes style and changeable attributes depending on the situation such as pose and illumination. In the second stage, the attribute code is divided into three types of codes once more as $\bm{a}_i = \left[ \bm{a}_i^{s} \, ; \bm{a}_i^{c} \, ; \bm{a}_i^{p}\right]$, which includes a style attribute code $\bm{a}_i^{s}$, an illumination attribute code $\bm{a}_i^{c}$, and a pose attribute code $\bm{a}_i^{p}$. The illumination and pose attribute codes $\bm{a}_i^{c}$ and $\bm{a}_i^{p}$ correspond to the cross-modality variation and the intra-modality variation, respectively. 
Note that we refer to the visual difference caused by different RGB and IR cameras as the illumination attribute. For clarity, both codes $\bm{a}_i^{c}$ and $\bm{a}_i^{p}$ can be integrated into an ID-excluded attribute code as $\bm{a}_i^{ex} = \left[ \bm{a}_i^{c} \, ; \bm{a}_i^{p}\right]$. In summary, ID-excluded factors involves the illumination attribute code $\bm{a}_i^{c}$ and the pose attribute code $\bm{a}_i^{p}$, while ID-discriminative factors corresponds to the style attribute code $\bm{a}_i^{s}$ and the prototype code $\bm{p}_i$. This assumption is different from the case of single-modality re-identification where color information is the key clue as discussed in \cite{disentangle_cvpr2019_DGNET}. 



\vspace{+0.5mm}
\textbf{Disentangling ID-excluded factors.} In the image generation process, our primary strategy is to synthesize a pair of cross-modality images by swapping the ID-excluded factors of two images with the same ID. Since two cross-modality images share the same ID characteristic, we can apply the image reconstruction loss between the translated image and the cross-modality image. Formally, this cross-modality reconstruction loss is formulated as follows:
\begin{equation}
\mathcal{L}^{cross}_{recon1} = \, 
\mathbb{E}_{
\renewcommand\arraystretch{0.4}
\begin{matrix}
\scriptscriptstyle{\bm{x}_1 \sim p_{d\!a\!t\!a}\!(\bm{x}_1),} 
\\ \scriptscriptstyle{\bm{x}_2 \sim p_{d\!a\!t\!a}\!(\bm{x}_2)}
\end{matrix}
}\!\left[\left \| \bm{x}_1 - G (\bm{p}_2, \bm{a}_2^{s}, \bm{a}_1^{ex}) \right \|_1\right],
\label{eq:cross_recon}
\end{equation}
\noindent
where $\bm{p}_i=E^p_i(\bm{x}_i)$, $\left[ \bm{a}_i^{s} \, ; \bm{a}_i^{ex}\right] = E^a_i(\bm{x}_i)$, and $G$ denotes a decoder. 
The $l_1$ loss encourages the generator to create sharp images. From this cross-modality reconstruction loss, the generator learns how to encode and decode the ID-excluded factors. To be clear, we only represent the loss for one modality as $\mathcal{L}^{cross}_{recon1}$. Another loss $\mathcal{L}^{cross}_{recon2}$ is defined by changing the index of modalities.

\begin{figure}[t]
\begin{center}
\includegraphics[width=1.0\linewidth]{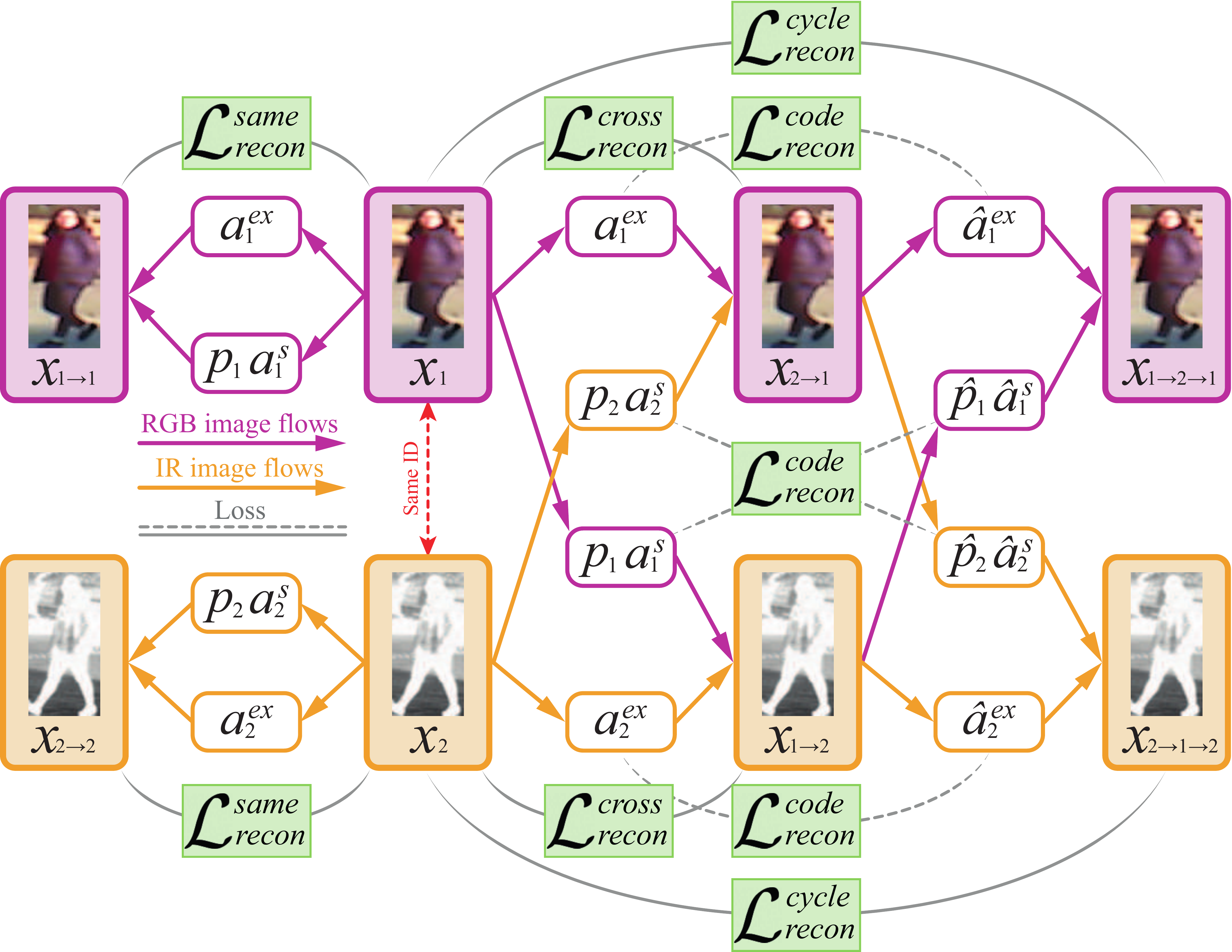}
\end{center}
\vspace{-2.5mm}
\caption{Illustration of the proposed reconstruction losses.}
\vspace{-2.5mm}
\label{fig:loss}
\end{figure}

\vspace{+0.2mm}
\textbf{Reconstruction loss.} We propose three additional reconstruction losses to improve the generation quality further, as expressed in Fig. \ref{fig:loss}. In addition to the loss of reconstructing images of different modalities, we apply a loss to reconstruct images of the same modality. This same-modality reconstruction loss plays a key role in regularization in the generation network, which is formulated as
\begin{equation}
\mathcal{L}^{same}_{recon1} = \, 
\mathbb{E}_{
\bm{x}_1 \sim p_{d\!a\!t\!a}\!(\bm{x}_1)
}\!\left[\left \| \bm{x}_1 - G (\bm{p}_1, \bm{a}_1^{s}, \bm{a}_1^{ex}) \right \|_1\right],
\label{eq:same_recon}
\end{equation}
\noindent
where $\bm{p}_i=E^p_i(\bm{x}_i)$ and $\left[ \bm{a}_i^{s} \, ; \bm{a}_i^{ex}\right] = E^a_i(\bm{x}_i)$.


We also apply the cycle consistency to regularize the ill-posed unsupervised image-to-image translation problem \cite{cycleGAN}. The cycle reconstruction loss is formulated as follows:
\begin{equation} 
\mathcal{L}^{cycle}_{recon1} = \, 
\mathbb{E}_{
\renewcommand\arraystretch{0.4}
\begin{matrix}
\scriptscriptstyle{\bm{x}_1 \sim p_{d\!a\!t\!a}\!(\bm{x}_1),} 
\\ \scriptscriptstyle{\bm{x}_2 \sim p_{d\!a\!t\!a}\!(\bm{x}_2)}
\end{matrix}
}\!\left[\left \| \bm{x}_1 - G (\hat{\bm{p}}_1, \hat{\bm{a}}_1^{s}, \hat{\bm{a}}_1^{ex}) \right \|_1\right],
\label{eq:cycle_recon}
\end{equation}
where  $\hat{\bm{p}}_1$,  $\hat{\bm{a}}_1^{s}$, and $\hat{\bm{a}}_1^{ex}$ denote the reconstructed prototype code, the reconstructed style attribute code, and the reconstructed ID-excluded attribute code, respectively. 
$\hat{\bm{p}}_1$, $\hat{\bm{a}}^{s}_1$, and $\hat{\bm{a}}^{ex}_1$ are obtained from $E_2^p( G (\bm{p}_1, \bm{a}_1^{s}, \bm{a}_2^{ex}) )$, $E_2^a( G (\bm{p}_1, \bm{a}_1^{s}, \bm{a}_2^{ex}) )$, and $E_1^a(G (\bm{p}_2, \bm{a}_2^{s}, \bm{a}_1^{ex}))$ respectively, where $\bm{p}_i=E^p_i(\bm{x}_i)$ and $\left[ \bm{a}_i^{s} \, ; \bm{a}_i^{ex}\right] = E^a_i(\bm{x}_i)$. 

Besides, we apply a code reconstruction loss as follows:
\begin{equation}
\begin{split}
\mathcal{L}^{code}_{recon1} = \, 
&\mathbb{E}_{
\renewcommand\arraystretch{0.4}
\begin{matrix}
\scriptscriptstyle{\bm{x}_1 \sim p_{d\!a\!t\!a}\!(\bm{x}_1),} 
\\ \scriptscriptstyle{\bm{x}_2 \sim p_{d\!a\!t\!a}\!(\bm{x}_2)}
\end{matrix}
} \left[\left \| \bm{a}_1^{s} - \hat{\bm{a}}_1^{s} \right \|_1\right]
\\
+&\mathbb{E}_{
\renewcommand\arraystretch{0.4}
\begin{matrix}
\scriptscriptstyle{\bm{x}_1 \sim p_{d\!a\!t\!a}\!(\bm{x}_1),} 
\\ \scriptscriptstyle{\bm{x}_2 \sim p_{d\!a\!t\!a}\!(\bm{x}_2)}
\end{matrix}
} \left[\left \| \bm{a}_1^{ex} - \hat{\bm{a}}_1^{ex} \right \|_1\right].
\end{split}
\label{eq:code_recon}
\end{equation}
\noindent
$\hat{\bm{a}}^{s}_1$ and $\hat{\bm{a}}^{ex}_1$ are obtained from $E_2^a( G (\bm{p}_1, \bm{a}_1^{s}, \bm{a}_2^{ex}) )$ and $E_1^a(G (\bm{p}_2, \bm{a}_2^{s}, \bm{a}_1^{ex}))$, where $\bm{p}_i=E^p_i(\bm{x}_i)$ and $\left[ \bm{a}_i^{s} \, ; \bm{a}_i^{ex}\right] = E^a_i(\bm{x}_i)$. This loss includes both assumptions that ID-discriminative factors should be preserved during the cross-modality reconstruction process as well as ID-excluded factors should be maintained during the same modality reconstruction process. The overall losses for reconstruction are expressed as follows:
\begin{equation} 
\mathcal{L}^{recon} = \lambda_{1} \mathcal{L}^{cross}_{recon} + \lambda_{2}  \mathcal{L}^{same}_{recon} + \lambda_{3} \mathcal{L}^{cycle}_{recon} + \lambda_{4} \mathcal{L}^{code}_{recon},
\label{eq:all_recon}
\end{equation}
\noindent
where $\lambda_t, t\in \{1,2,3,4\}$ controls the relative importance of four losses. $\mathcal{L}^{cross}_{recon}$ indicates the sum of $ \mathcal{L}^{cross}_{recon1}$ and $\mathcal{L}^{cross}_{recon2}$. Other losses $\mathcal{L}^{same}_{recon}$, $\mathcal{L}^{cycle}_{recon}$, $\mathcal{L}^{code}_{recon}$ are calculated in the same manner.

\vspace{+0.5mm}
\textbf{KL divergence loss.} To help the attribute encoder $E_a$ learn more informative representations, we utilize the Kullback-Leibler (KL) divergence loss. This loss encourages the ID-excluded attribute representation to be as close to a prior Gaussian distribution as follows:
\begin{equation} \
\mathcal{L}^{kl}_{1} = \,
\mathbb{E}_{\bm{x}_1 \sim p(\bm{x}_1)}\left[ \mathcal{D}_{KL}(\bm{a}_1^{ex} \| N(0, 1)) \right],
\label{eq:KLD}
\end{equation}
\noindent
where $\mathcal{D}_{KL}(p\|q) = - \int p(z) \log{\frac{p(z)}{q(z)}dz}$ and $\left[ \bm{a}_1^{s} \, ; \bm{a}_1^{ex}\right] = E^a_1(\bm{x}_1)$. By limiting the distribution range of the cross-modality and intra-modality characteristics, this KL divergence loss enables ID-excluded attribute codes to change continuously in the latent space. $\mathcal{L}^{kl}_2$ is defined in a similar manner and $\mathcal{L}^{kl} = \mathcal{L}^{kl}_{1} + \mathcal{L}^{kl}_{2}$.

\vspace{+0.5mm}
\textbf{Adversarial loss.} Since generating realistic images is crucial for image-to-image translation, we apply an adversarial loss \cite{GAN} by using the cross-reconstructed images with different modalities. Two discriminators $D_1, D_2$ corresponding visible and infrared domains are employed for adversarial training. In the case of modality 1, the RGB discriminator $D_1$ distinguishes the real image $\bm{x}_1$ and the fake image $G (\bm{p}_2, \bm{a}_2^{s}, \bm{a}_1^{ex})$ used for cross-modality reconstruction. The generator tries to synthesize a more realistic RGB image to fool the discriminator. Accordingly, the ID-excluded attribute code $\bm{a}_1^{ex}$ is encouraged to include the modality characteristic of RGB.

Furthermore, we introduce a new strategy to distinguish the cross-modality characteristic (\ie illumination) and the intra-modality characteristic (\ie pose). As mentioned above, the ID-excluded attribute code can be divided into two attribute codes as $\bm{a}^{ex}_1 = \left[ \bm{a}^c_1 \,; \bm{a}^p_1\right]$. Our idea is to swap only the illumination attribute leaving the pose attribute unchanged as $G (\bm{p}_2, \bm{a}_2^{s}, \bm{a}_1^{c}, \bm{a}_2^{p})$. By feeding this image to the RGB discriminator, the modality characteristic is concentrated only on the illumination attribute code $\bm{a}_1^{c}$. The remaining intra-modality characteristic across RGB-IR images is collected in the pose attribute code $\bm{a}_2^{p}$. Adversarial losses are employed to play the minimax game, which is formulated as follows:
\vspace{-1.0mm}
\begin{equation} 
\begin{split}
\mathcal{L}^{adv}_{1} = \, 
&\mathbb{E}\!_{
\renewcommand\arraystretch{0.4}
\begin{matrix}
\scriptscriptstyle{\bm{x}_1 \sim p_{d\!a\!t\!a}\!(\bm{x}_1),} 
\\ \scriptscriptstyle{\bm{x}_2 \sim p_{d\!a\!t\!a}\!(\bm{x}_2)}
\end{matrix}
}\!\left[\log{(1\!-\!D_1 (G (\bm{p}_2, \bm{a}_2^{s}, \bm{a}_1^{c}, \bm{a}_1^{p})))}\right]
\\
+&\mathbb{E}\!_{
\renewcommand\arraystretch{0.4}
\begin{matrix}
\scriptscriptstyle{\bm{x}_1 \sim p_{d\!a\!t\!a}\!(\bm{x}_1),} 
\\ \scriptscriptstyle{\bm{x}_2 \sim p_{d\!a\!t\!a}\!(\bm{x}_2)}
\end{matrix}
}\!\left[\log{(1\!-\!D_1 (G (\bm{p}_2, \bm{a}_2^{s}, \bm{a}_1^{c}, \bm{a}_2^{p})))}\right]
\\
+&\mathbb{E}_{\bm{x}_1 \sim p(\bm{x}_1)}\left[ \log{D_1(\bm{x}_1)} \right],
\end{split}
\vspace{-0.0mm}
\label{eq:adv}
\end{equation}
\noindent
where $\bm{p}_i=E^p_i(\bm{x}_i)$ and $\left[ \bm{a}_i^{s} \, ; \bm{a}_i^{c} \, ;  \bm{a}_i^{p}\right] = E^a_i(\bm{x}_i)$.
The generator is trained to minimize (\ref{eq:adv}) while the discriminator attempts to maximize it. 
Especially, the parameters of the discriminator are updated when the parameters of the generator are fixed.
$\mathcal{L}^{adv}_{2}$ is defined in a similar way and $\mathcal{L}^{adv} = \mathcal{L}^{adv}_{1} + \mathcal{L}^{adv}_{2}$.

\subsection{Hierarchical Feature Learning} \label{sec:HFL}

As illustrated in Fig. \ref{fig:framework}, our Hierarchical Feature Learning (HFL) module is coupled with ID-PIG by sharing the prototype and attribute encoders. This module enables both encoders to extract the common ID-discriminative factors between RGB-IR images. At the same time, this feature learning process implicitly assists in separating intra- and cross-modality characteristics from cross-modality images and enhances the quality of image generation. 

\vspace{+0.5mm}
\textbf{Re-entangling ID-discriminative factors.} We introduce the ID-discriminative feature by concatenating the style attribute code and the prototype code to distinguish person identities. Compared to using one of the two codes, the combination of both codes with different characteristics encourages the network to learn rich representations of a person's identity. Given the prototype tensor $\bm{p}_i$ from the prototype encoder $E^p_i$, the feature embedding network $H$ projects it to the ID-discriminative prototype code $\bm{p}^{d}_i $, where $\bm{p}^{d}_i = H(\bm{p}_i)$. We then concatenate the ID-discriminative prototype code $\bm{p}^{d}_i $ and the style attribute code $\bm{a}^{s}_i $ with a learnable parameter $\alpha \in [0, 1]$, which is expressed as $\bm{d}^{comb}_i\!=\!\left[  \alpha\!\cdot\!\bm{p}^{d}_i ; (1-\alpha)\!\cdot\!\bm{a}^{s}_i  \right]$. Then, the combined code $\bm{d}^{comb}_i$ is feed into a fully connected layer. In the testing phase, we use the output $\bm{f}$ of the fully connected layer for cross-modality retrieval with the Euclidean distance.


\vspace{+0.5mm}
\textbf{Alternate sampling strategy.} We form a set of training feature vectors $\bm{f} \in \mathcal{F}$ by selecting various types of style attribute codes and prototype codes alternately. This alternate sampling strategy improves the discrimination ability by overcoming the lack of diversity in the training dataset. We alternately combine style attribute codes $\bm{a}^{s}$ and prototype codes $\bm{p}^{d}$ extracted from the original images $\bm{x}_1$, $\bm{x}_2$ and the cross-reconstructed images $\bm{x}_{1 \rightarrow 2} = G (\bm{p}_1, \bm{a}_1^{s}, \bm{a}_2^{c}, \bm{a}_2^{p})$, $\bm{x}_{2 \rightarrow 1} = G (\bm{p}_2, \bm{a}_2^{s}, \bm{a}_1^{c}, \bm{a}_1^{p})$. Note that the attribute and prototype codes for combination must be of the same person. 



\vspace{+0.5mm}
\textbf{Cross-entropy loss.} Given a set of training feature vectors with the identity annotation $\left\{\bm{f}_i, y_i\right\}$, we use the cross-entropy loss for ID-discriminative learning, which is expressed as follows:
\begin{equation} 
\mathcal{L}^{ce} = \,
\mathbb{E}_{\bm{f} \in \mathcal{F}, y \sim \mathcal{Y}}\left[ - \log({p(y| \bm{f})}) \right],
\label{eq:CE}
\end{equation}
\noindent
where $p(y| \bm{f})$ indicates the predicted probability of a sampled feature vector $\bm{f}$ belonging to the identity $y$. 


\vspace{+0.5mm}
\textbf{Triplet loss.} For similarity learning, we also employ the triplet loss. The triplet loss is expressed as follows:
\vspace{-0.5mm}
\begin{equation} 
\mathcal{L}^{trip} = \,
\sum_{\bm{f}^a, \bm{f}^p, \bm{f}^n \in \mathcal{F}} \left[  d(\bm{f}^a, \bm{f}^p) - d(\bm{f}^a, \bm{f}^n) + m \right]_+,
\vspace{-0.5mm}
\label{eq:TRIP}
\end{equation}
\noindent
where $\bm{f}^a$, $\bm{f}^p$, and $\bm{f}^n$ indicate anchor, positive, and negative samples. $d(\cdot, \cdot)$ is the Euclidean distance, $m$ is a margin parameter, and $[z]_+ = \max(z, 0)$. For each sample $\bm{f}^a$ in the set $\mathcal{F}$, we select the hardest positive sample $\bm{f}^p$  and the hardest negative samples $\bm{f}^n$ within the batch in the same way as \cite{TRIP}. The triplet loss forces intra-class samples closer and inter-class samples farther. As a result, the cross-entropy and triplet losses help the encoder to clearly disentangle ID-discriminative factors and ID-excluded factors from RGB-IR images.

\textbf{End-to-end training.} As a summary, the overall loss for our Hi-CMD method is expressed as follows:
\vspace{-0.5mm}
\begin{equation} 
\mathcal{L} = \mathcal{L}^{recon} + \lambda_{kl}\mathcal{L}^{kl} + \lambda_{adv}\mathcal{L}^{adv} + \lambda_{ce}\mathcal{L}^{ce}  + \lambda_{trip}\mathcal{L}^{trip},
\vspace{-0.5mm}
\label{eq:total_loss}
\end{equation}
\noindent
where $\lambda_{kl}$, $\lambda_{adv}$, $\lambda_{ce}$, and $\lambda_{trip}$ are hyperparameters to control the relative importance of loss terms. We train the whole network to optimize the total loss in an end-to-end manner. For adversarial learning, we alternatively train the parameters of discriminators and the remaining parameters.

\subsection{Discussion}
\vspace{-0.5mm}
We compare Hi-CMD with the most related disentanglement approach DG-Net \cite{disentangle_cvpr2019_DGNET}. DG-Net is similar to our proposed Hi-CMD in that both methods combine an image generation network with a discriminative learning module in an end-to-end manner. However, the disentangled elements and the feature vectors used for person ReID are entirely different. 
While DG-Net decomposes each RGB image into appearance and structure codes, our Hi-CMD hierarchically disentangles ID-discriminative factors and ID-excluded factors including pose and illumination attributes from RGB-IR images. 
In addition, DG-Net only uses an appearance code where color information is vital to distinguish people.
However, since this factor is not feasible in the VI-ReID task, we manage the ID-discriminative information between RGB-IR images by hierarchical disentanglement. This hierarchical approach is more useful for extracting the common ID-discriminative feature. 



%
%
%
%





\begin{table}[] 
\renewcommand*{\arraystretch}{1.0}
\footnotesize
\centering
\begin{tabular}{M{2.20cm}||M{0.46cm}M{0.46cm}M{0.6cm}||M{0.46cm}M{0.46cm}M{0.55cm}}
\hline
Datasets                    & \multicolumn{3}{c||}{RegDB \cite{dataset_RegDB}}    & \multicolumn{3}{c}{SYSU-MM01 \cite{vireid_iccv2017_SYSU}} \\ \hline
Methods                     & $R$=1   & $R$=10  & mAP   & $R$=1  & $R$=10      & mAP         \\ \hhline{=||===||===}
HOG \cite{HOG}   & 13.49 & 33.22 & 10.31 & 2.76         & 18.25          & 4.24        \\
LOMO \cite{LOMO} & 0.85  & 2.47   & 2.28  & 1.75         & 14.14            & 3.48        \\
MLBP \cite{MLBP} & 2.02  & 7.33   & 6.77  & 2.12         & 16.23          & 3.86        \\
GSM \cite{GSM} & 17.28 & 34.47  & 15.06 & 5.29         & 33.71            & 8.00        \\
SVDNet \cite{SVDNET}  & 17.24 & 34.12  & 19.04 & 14.64        & 53.28            & 15.17       \\
PCB \cite{PCB}  & 18.32 & 36.42  & 20.13 & 16.43        & 54.06         & 16.26       \\ \hline
One stream \cite{vireid_iccv2017_SYSU} & 13.11 & 32.98  & 14.02 & 12.04        & 49.68           & 13.67       \\
Two stream \cite{vireid_iccv2017_SYSU} & 12.43 & 30.36  & 13.42 & 11.65        & 47.99           & 12.85       \\
Zero padding \cite{vireid_iccv2017_SYSU} & 17.75 & 34.21  & 18.90 & 14.80        & 54.12         & 15.95       \\
TONE \cite{vireid_aaai2018_TONE} & 16.87 & 34.03  & 14.92 & 12.52        & 50.72        & 14.42       \\
TONE\scriptsize{+}\footnotesize{HCML}\cite{vireid_aaai2018_TONE} & 24.44 & 47.53 & 20.80 & 14.32        & 53.16        & 16.16       \\
BCTR \cite{vireid_ijcai2018_BCTR} & 32.67 & 57.64  & 30.99 & 16.12        & 54.90           & 19.15       \\
BDTR \cite{vireid_ijcai2018_BCTR} & 33.47 & 58.42  & 31.83 & 17.01        & 55.43         & 19.66       \\
eBDTR\scriptsize(alex) \footnotesize\cite{vireid_tifs2019_eBDTR} & 34.62 & 58.96  & 33.46 & 22.42        & 64.61        & 24.11       \\
eBDTR\scriptsize(resnet) \footnotesize\cite{vireid_tifs2019_eBDTR} & 31.83 & 56.12  & 33.18 & 27.82        & 67.34       & 28.42       \\
cmGAN \cite{vireid_ijcai2018_cmGAN} & -     & -     & -       & 26.97        & 67.51           & 27.80       \\
D2RL  \cite{vireid_cvpr2019_D2RL}  & 43.40 & 66.10  & 44.10 & {\color{blue}\textbf{28.90}}        & {\color{blue}\textbf{70.60}}   & {\color{blue}\textbf{29.20}}       \\
HSME  \cite{vireid_aaai2019_HSME}  & 41.34 & 65.21  & 38.82 & 18.03        & 58.31       & 19.98       \\
D-HSME \cite{vireid_aaai2019_HSME} & {\color{blue}\textbf{50.85}} & {\color{blue}\textbf{73.36}} & {\color{blue}\textbf{47.00}} & 20.68        & 62.74          & 23.12       \\ \hhline{=||===||===}
Ours (Hi-CMD)             & {\color{red}\textbf{70.93}}   & {\color{red}\textbf{86.39}}   &  {\color{red}\textbf{66.04}}   &  {\color{red}\textbf{34.94}}  & {\color{red}\textbf{77.58}}   &     {\color{red}\textbf{35.94}}   \\ \hline
\end{tabular}
\vspace{0mm}
\caption{Comparison with the state-of-the-arts on RegDB and SYSU-MM01 datasets. Re-identification rates (\%) at rank $R$ and mAP (\%). $1^{st}$ and $2^{nd}$ best results are indicated by {\color{red}\textbf{red}} and {\color{blue}\textbf{blue}} color, respectively.}
\vspace{-3mm}
\label{table:all}
\end{table}

\section{Experiments}
\vspace{-0.5mm}
\subsection{Datasets and Settings}
\vspace{-0.5mm}
\textbf{Datasets.} Extensive experiments were conducted on two widely used VI-ReID datasets, RegDB \cite{dataset_RegDB} and SYSU-MM01 \cite{vireid_iccv2017_SYSU}. We followed the RegDB evaluation protocol in \cite{vireid_aaai2018_TONE, vireid_ijcai2018_BCTR} and the SYSU-MM01 evaluation protocol in \cite{vireid_iccv2017_SYSU}. The RegDB dataset consists of 2,060 visible images and 2,060 far-infrared images with 206 identities for training. The testing set contains 206 identities with 2,060 visible images for the query and 2,060 far-infrared images for the gallery. We repeated 10 trials with a random split to achieve statistically stable results. The SYSU dataset contains 22,258 visible images and 11,909 near-infrared images of 395 identities for training. The testing set includes 96 identities with 3,803 near-infrared images for the query and 301 visible images as the gallery set. The SYSU dataset is collected by six cameras (four visible and two near-infrared), including indoor and outdoor environments. We adopted the most challenging \textit{single-shot all-search} mode and repeated the above evaluation 10 trials with a random split of the gallery and probe set. 



\textbf{Evaluation metrics.} Two popular evaluation metrics are adopted: Cumulative Matching Characteristic (CMC) and mean Average Precision (mAP). The rank-$k$ identification rate in the CMC curve indicates the cumulative rate of true matches in the top-$k$ position. The other evaluation metric is the mean average precision (mAP), considering person re-identification as a retrieval task.

\textbf{Implementation details.} Our method is implemented with the Pytorch framework on an NVIDIA Titan Xp GPU. Visible and infrared images are resized to $256 \times 128 \times 3$. Each mini-batch contains 4 pairs of visible and infrared images with different identities. The reconstruction parameters $\lambda_1, \lambda_2, \lambda_3, \lambda_4$ in (\ref{eq:all_recon}) were set to 50, 50, 50, 10, respectively. The parameters $\lambda_{kl}, \lambda_{adv}, \lambda_{ce}, \lambda_{trip}$ in (\ref{eq:total_loss}) were set to 1, 20, 1, 1, respectively. We used Stochastic Gradient Descent with the learning rate 0.001 and momentum 0.9 to optimize the HFL module. We adopted the Adam optimizer \cite{ADAM} with the learning rate 0.0001 for the ID-PIG network. The ID-PIG framework is modified based on MUNIT \cite{disentangle_eccv2018_MUNIT} and the feature embedding network $H$ is based on ResNet-50 \cite{RESNET} pretrained on ImageNet \cite{IMAGENET}. For more details, please refer to the supplementary material.

\subsection{Comparison with State-of-the-art Methods}

\begin{figure}[t]
\begin{center}
\includegraphics[width=1.0\linewidth]{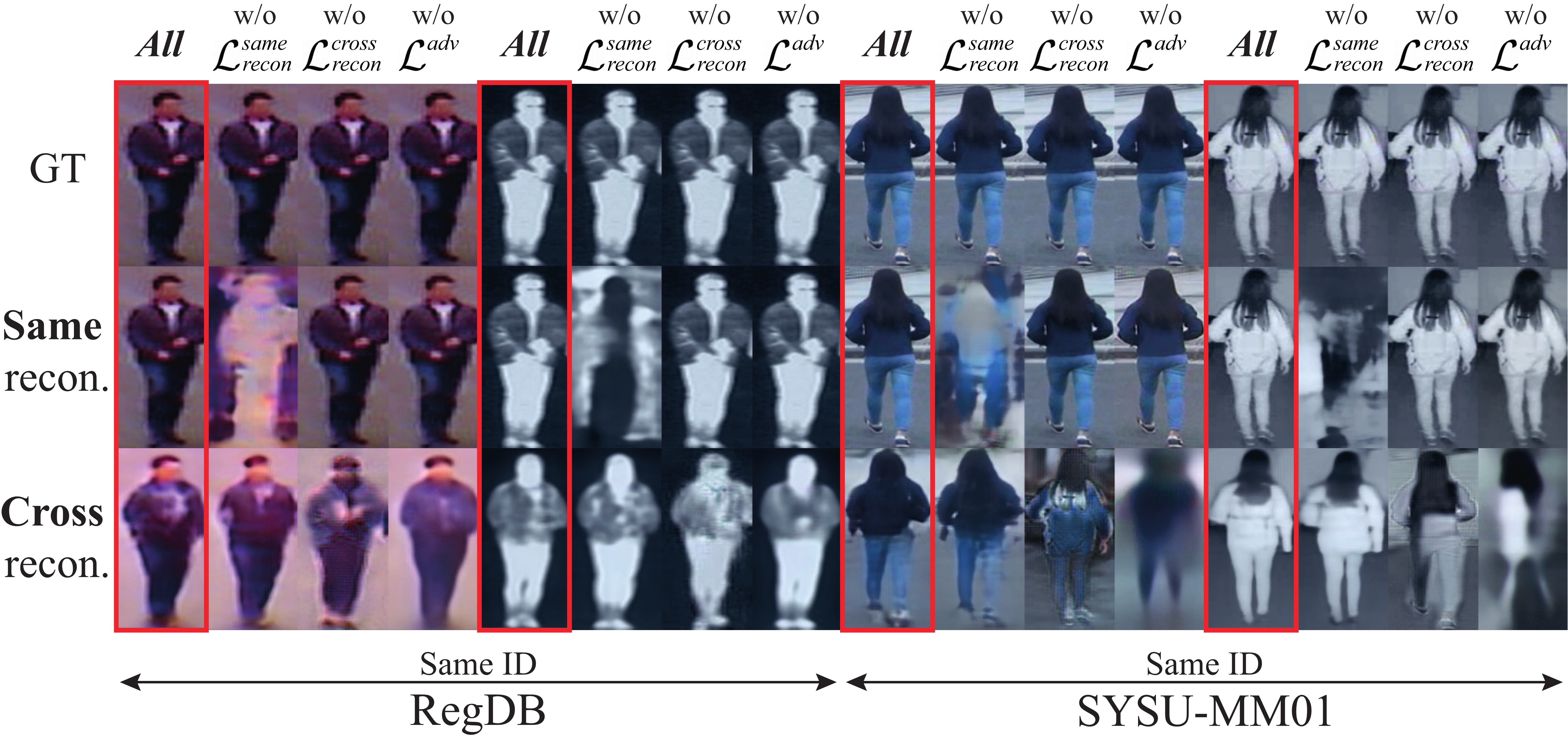}
\end{center}
\vspace{-3mm}
\caption{Qualitative comparison between image generation networks for various loss combinations on RegDB and SYSU-MM01. Zoom in for best view.}
\vspace{-3mm}
\label{fig:I2I}
\end{figure}

\textbf{Comparison with conventional methods.} 
We compare our method with conventional methods, which are not designed for VI-ReID. Feature extraction (HOG \cite{HOG}, LOMO \cite{LOMO}, MLBP \cite{MLBP}), cross-domain matching (GSM \cite{GSM}), and RGB-based person ReID (SVDNET \cite{SVDNET}, PCB \cite{PCB}) methods are included for comparison. Table \ref{table:all} shows that all methods have relatively poor performance. Although the PCB method achieves high performance in single-modality person ReID, a significant performance drop is inevitable in the VI-ReID task. Note that a pixel-level difference between visible and infrared images is challenging to deal with at the feature-level representation.


\textbf{Comparison with state-of-the-arts.} 
We compare our method with the state-of-the-art methods in VI-ReID. The competing methods include feature learning frameworks (one-stream, two-stream, zero-padding \cite{vireid_iccv2017_SYSU}, and TONE \cite{vireid_aaai2018_TONE}), ranking losses (BCTR \cite{vireid_ijcai2018_BCTR}, BDTR \cite{vireid_ijcai2018_BCTR}, eBDTR \cite{vireid_tifs2019_eBDTR}), metric learning (HCML \cite{vireid_aaai2018_TONE}, HSME \cite{vireid_aaai2019_HSME}, D-HSME \cite{vireid_aaai2019_HSME}), reducing distribution divergence (cmGAN \cite{vireid_ijcai2018_cmGAN}), and image generation (D2RL \cite{vireid_cvpr2019_D2RL}) methods. Our model achieves 70.93\% rank-1 identification rate and 66.04\% mAP score on the RegDB dataset \cite{dataset_RegDB}, and 34.94\% rank-1 identification rate and 35.94\% mAP score on the SYSU-MM01 dataset \cite{vireid_iccv2017_SYSU}. Our method significantly outperforms the state-of-the-art VI-ReID methods on both the RegDB and SYSU-MM01 datasets. This comparison indicates the effectiveness of our disentanglement approach for bridging the cross-modality and intra-modality gaps. Moreover, this improvement in performance can be analyzed by the visualization of ID-discriminative factors, which is discussed in Section \ref{sec:PAE}.


\subsection{Further Evaluations and Analysis}

\textbf{Impact of image generation losses.} We performed an ablation study of our ID-preserving Person Image Generation (ID-PIG) network. To evaluate image generation losses qualitatively, we compare four variations of ID-PIG: 1) our best model with all components; 2) removing the reconstruction loss $\mathcal{L}^{same}_{recon}$; 3) removing the disentanglement loss $\mathcal{L}^{cross}_{recon}$; 4) removing the adversarial loss $\mathcal{L}^{adv}$. The network structure and training strategy remain the same for all settings. Figure \ref{fig:I2I} shows the results of this image translation experiment. The samples are randomly selected from the testing set. We observe that the generated images can contain unpleasant artifacts such as blurriness or color shifts if one kind of loss is excluded from the training process. On the other hand, the results generated by our ID-PIG network with all components show more a realistic and sharp appearance regardless of the modalities are changed or not.

\begin{figure}[t]
\begin{center}
\includegraphics[width=1.0\linewidth]{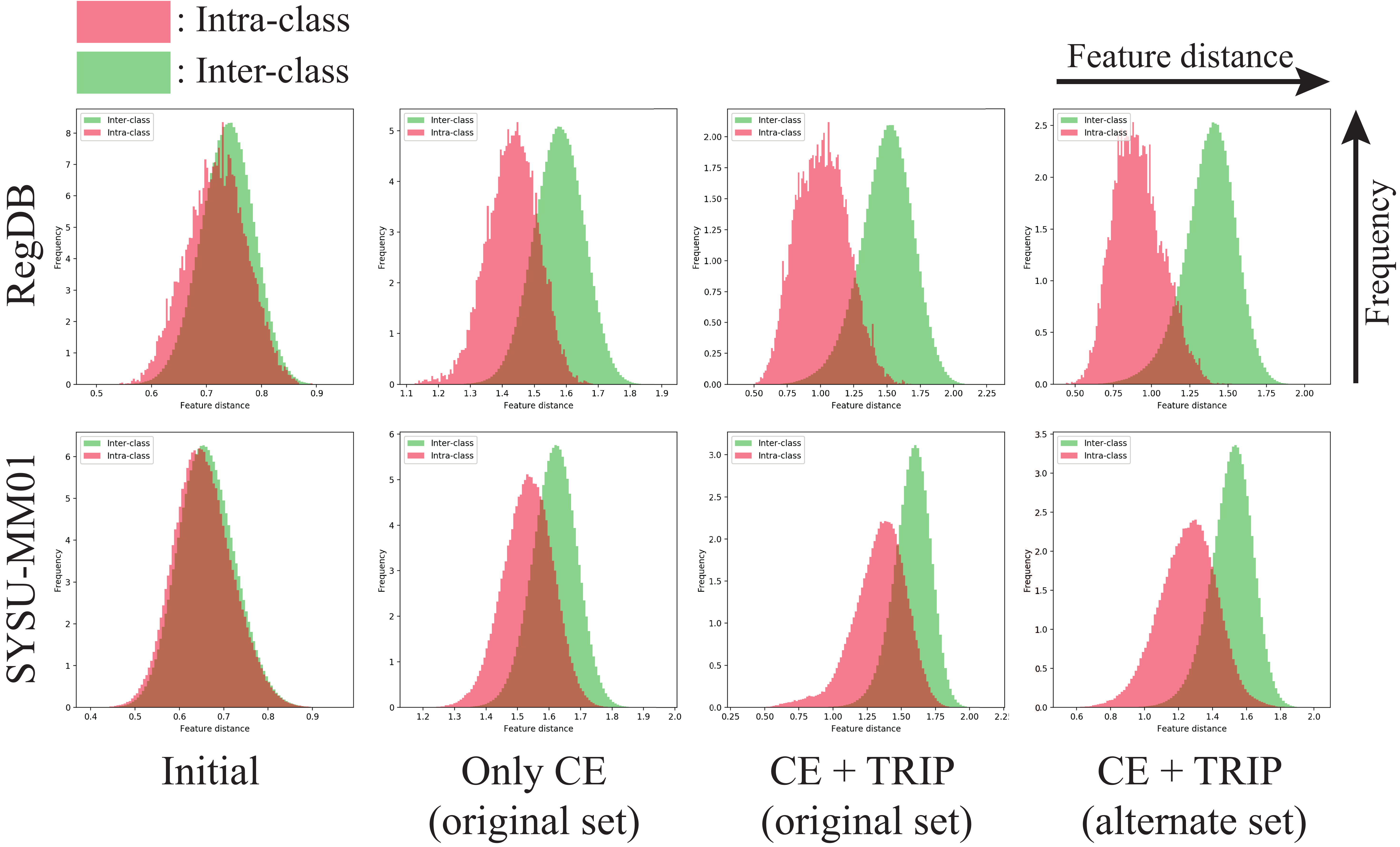}
\end{center}
\vspace{-3mm}
\caption{The distribution of the Euclidean distance between cross-modality (RGB-IR) features. The intra-class and inter-class distances are indicated by \textcolor{myred}{\textbf{red}} and \textcolor{mygreen}{\textbf{green}} color, respectively.}
\vspace{-2mm}
\label{fig:HFL}
\end{figure}

\begin{table}[]
\renewcommand*{\arraystretch}{1.0}
\footnotesize
\centering
\begin{tabular}{M{1.0cm}|M{1.05cm}|M{0.75cm}||M{0.6cm}M{0.6cm}||M{0.6cm}M{0.6cm}}
\hline
\multicolumn{3}{c||}{Methods}     & \multicolumn{2}{c||}{RegDB} & \multicolumn{2}{c}{SYSU-MM01} \\ \hline
Input set     & Loss    & Feature & R=1          & mAP         & R=1           & mAP           \\ \hhline{=|=|=||==||==}
Original  & CE      & A+P    & 36.36        & 33.47       & 18.65         & 19.49        \\
Original  & CE+TRIP & A      & 15.33        & 15.37       & 6.05         & 7.74         \\
Original  & CE+TRIP & P      & 49.02        & 45.75       & 22.51         & 23.73         \\
Original  & CE+TRIP & A+P    & 53.25        & 49.53       & 29.19         & 30.53         \\  \hhline{=|=|=||==||==}
Alternate & CE+TRIP & A+P    & \textbf{70.93}        & \textbf{66.04}      & \textbf{34.94}         & \textbf{35.94}         \\ \hline
\end{tabular}
\vspace{1mm}
\caption{Component analysis of our HFL module on RegDB and SYSU-MM01 datasets. $A$ and $P$ represent the style attribute code and the prototype code respectively, which are used to train HFL.}
\label{table:HFL}
\vspace{-2mm}
\end{table}

\textbf{Effectiveness of hierarchical feature learning.}
We study the several variants of the Hierarchical Feature Learning (HFL) module on both datasets to demonstrate the effectiveness of our hierarchical disentanglement approach. Figure \ref{fig:HFL} shows the distribution of the Euclidean distances between RGB-IR images from the testing set. Compared to the initial state, the HFL module based on an alternate sampling strategy minimizes the intra-class distance and maximizes the inter-class distance simultaneously. Moreover, as shown in Table \ref{table:HFL}, the use of the alternate sampling strategy significantly improves performance compared to learning with the original image set. Our HFL method also achieves much higher performance than learning the style attribute code or the prototype code alone. These results prove that diverse code combinations by an alternate sampling strategy improve the discrimination ability and reduce the cross-modality gap significantly.

\subsection{Person Attribute Exploration} \label{sec:PAE}
\vspace{-0.5mm}

\begin{figure}[t]
\begin{center}
\includegraphics[width=1.0\linewidth]{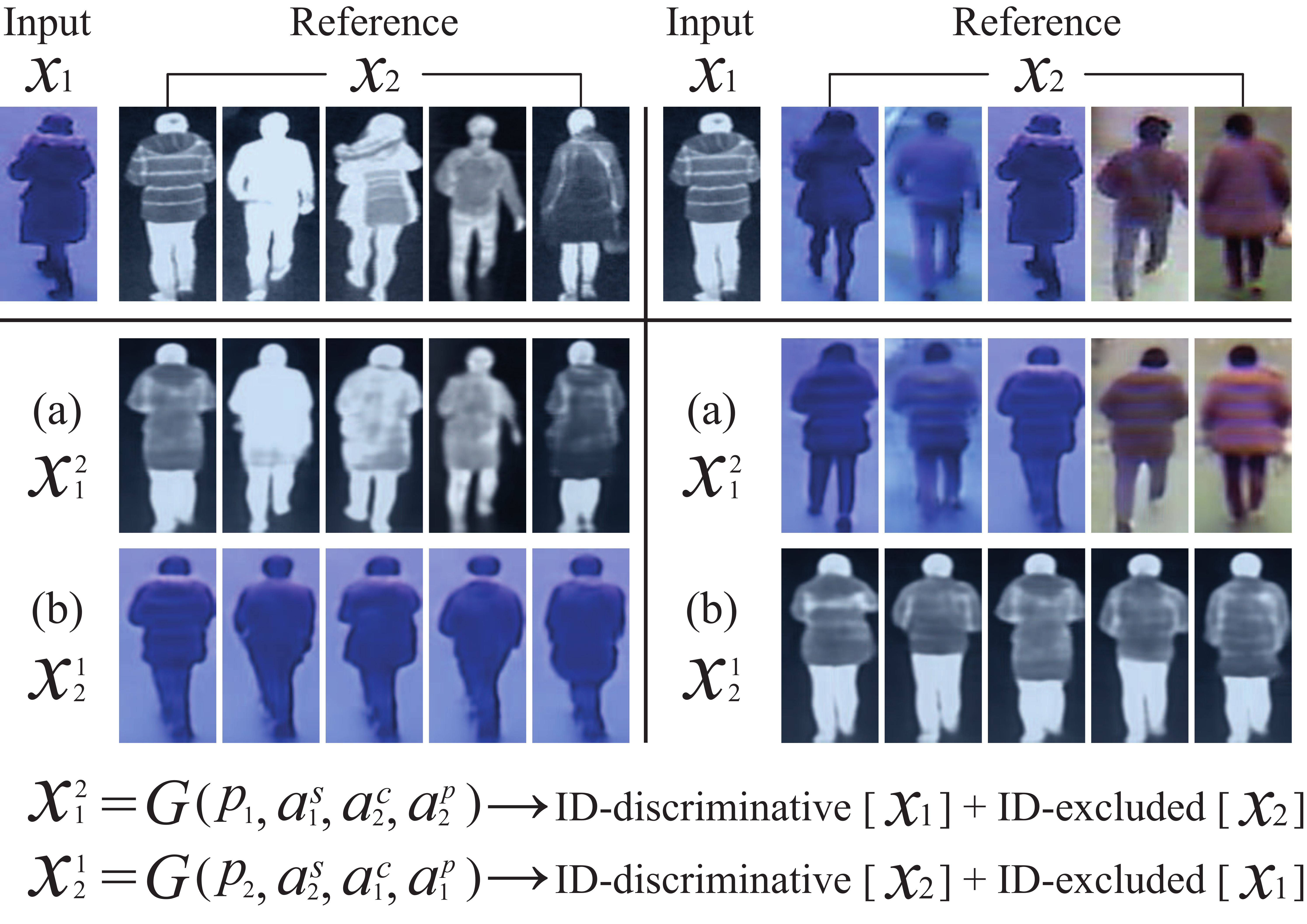}
\end{center}
\vspace{-3.5mm}
\caption{Examples of person image generation across two different swapping manners: (a) swapping ID-excluded factors; (b) swapping ID-discriminative factors.}
\vspace{-1mm}
\label{fig:ID_PIG1}
\end{figure}

\textbf{Disentangling ID-discriminative and ID-excluded factors.} In this section, we present image generation results by our ID-PIG network. To demonstrate that ID-discriminative factors and ID-excluded factors are clearly disentangled from RGB-IR images, we conducted two experiments: 1) swapping the ID-excluded factors of $\bm{x}_1$ and $\bm{x}_2$ with preserving ID-discriminative factors; 2) swapping the ID-discriminative factors of  $\bm{x}_1$ and $\bm{x}_2$ with maintaining ID-excluded factors. 
The images in Fig. \ref{fig:ID_PIG1} (a) are changed to the pose and illumination attributes of the reference images, while the clothes and patterns of the input image are retained. The images in Fig. \ref{fig:ID_PIG1} (b) are synthesized with the clothes of the reference images, while maintaining the pose and illumination attributes of the input image.
The visualization of ID-PIG proves that the shapes, patterns, and styles of clothes are important factors for cross-modality matching, which means that our novel hierarchical disentanglement approach can extract the common ID-discriminative feature effectively. 

%


\begin{figure}[t]
\begin{center}
\includegraphics[width=1.0\linewidth]{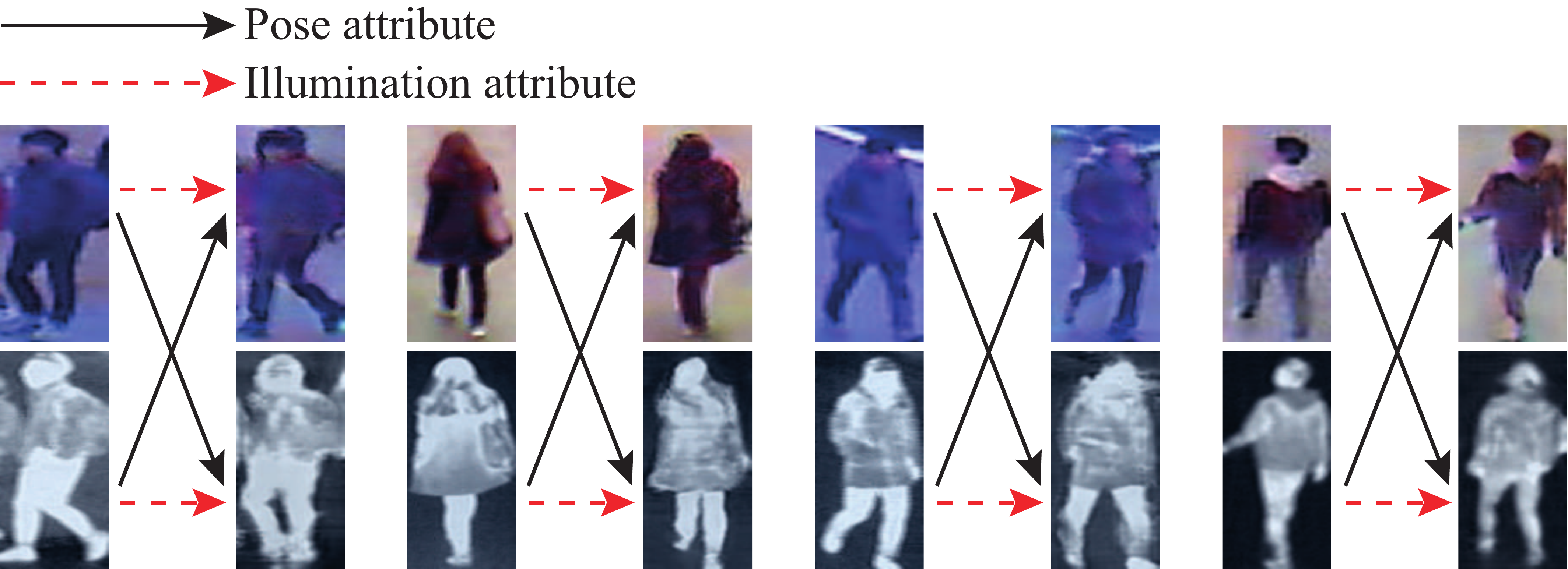}
\end{center}
\vspace{-4mm}
\caption{Examples of ID-excluded factor manipulation.}
\vspace{-2mm}
\label{fig:ID_PIG2}
\end{figure}

\textbf{Disentangling illumination and pose attributes.} To demonstrate that our proposed system can manipulate pose and illumination attributes in ID-excluded factors independently, we conducted an experiment that changes illumination attribute codes while maintaining pose attribute codes, as shown in Fig. \ref{fig:ID_PIG2}. Note that our adversarial loss enables the network to distinguish pose and illumination attributes in (\ref{eq:adv}). This experiment shows that the pose can be transformed into other poses without any supervision of human pose estimation, unlike conventional pose generation methods \cite{I2Ireid_cvpr2018_trans, I2Ireid_eccv2018_PNGAN}.

\begin{figure}[t]
\begin{center}
\includegraphics[width=1.0\linewidth]{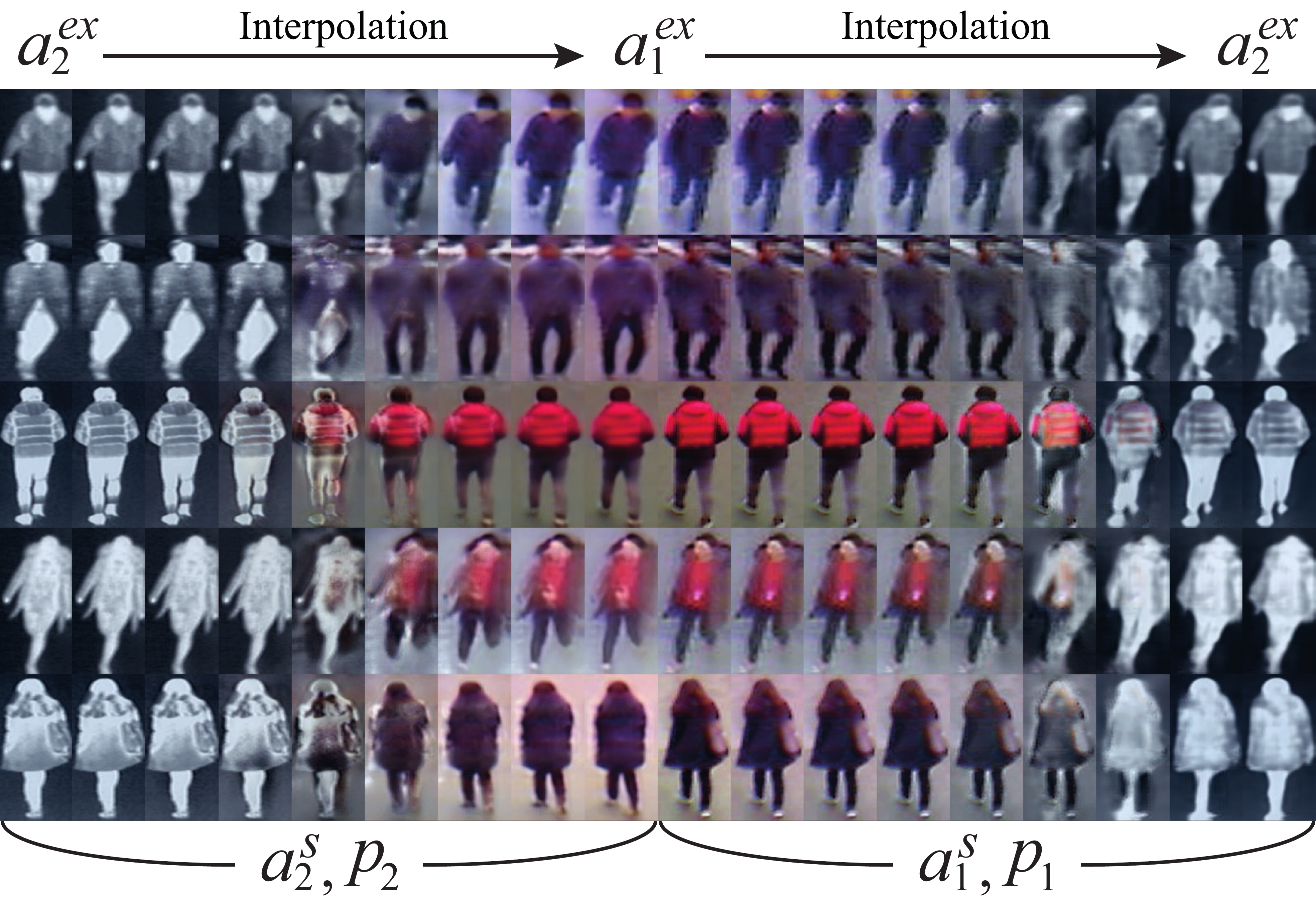}
\end{center}
\vspace{-4mm}
\caption{Examples of person image generation by linear interpolation of ID-excluded factors.}
\vspace{-1mm}
\label{fig:ID_PIG3}
\end{figure}

\textbf{Interpolation of ID-excluded attributes.} In this part, we validate that ID-excluded attributes of the generated images continuously change in the latent space. We extracted the ID-excluded attribute codes from a pair of images and obtained a series of attribute codes by linear interpolation. As shown in Fig. \ref{fig:ID_PIG3}, the pose and illumination are changed smoothly from left to right. This experiment ensures that our ID-PIG can synthesize a lot of unseen poses or illuminations.



\vspace{-0.5mm}
\section{Conclusion}
\vspace{-0.5mm}

In this work, we have proposed a novel hierarchical cross-modality disentanglement method for VI-ReID. In contrast to previous studies, the proposed model reduces both cross- and intra-modality discrepancies at the same time by disentangling ID-discriminative factors and ID-excluded factors from visible-infrared images. Furthermore, our novel ID-preserving person image generation network provides resources to analyze cross-modality matching results and alleviates the problem of the insufficient data. Quantitative and qualitative evaluation on VI-ReID datasets demonstrates the superiority of our proposed method over the state-of-the-art methods.


{\small
\bibliographystyle{ieee_fullname}
\bibliography{cvpr_bibtex}

\begin{thebibliography}{10}\itemsep=-1pt

\bibitem{disentangle_cvpr2018_preserving}
Jianmin Bao, Dong Chen, Fang Wen, Houqiang Li, and Gang Hua.
\newblock Towards open-set identity preserving face synthesis.
\newblock In {\em Proceedings of the IEEE Conference on Computer Vision and
  Pattern Recognition (CVPR)}, pages 6713--6722, 2018.

\bibitem{vireid_ijcai2018_cmGAN}
Pingyang Dai, Rongrong Ji, Haibin Wang, Qiong Wu, and Yuyu Huang.
\newblock Cross-modality person re-identification with generative adversarial
  training.
\newblock In {\em Proceedings of the International Joint Conference on
  Artificial Intelligence (IJCAI)}, pages 677--683, 2018.

\bibitem{HOG}
Navneet Dalal and Bill Triggs.
\newblock Histograms of oriented gradients for human detection.
\newblock 2005.

\bibitem{IMAGENET}
Jia Deng, Wei Dong, Richard Socher, Li-Jia Li, Kai Li, and Li Fei-Fei.
\newblock Imagenet: A large-scale hierarchical image database.
\newblock In {\em 2009 IEEE Conference on Computer Vision and Pattern
  Recognition (CVPR)}, pages 248--255. Ieee, 2009.

\bibitem{I2Ireid_cvpr2018_SPGAN}
Weijian Deng, Liang Zheng, Qixiang Ye, Guoliang Kang, Yi Yang, and Jianbin
  Jiao.
\newblock Image-image domain adaptation with preserved self-similarity and
  domain-dissimilarity for person re-identification.
\newblock In {\em Proceedings of the IEEE Conference on Computer Vision and
  Pattern Recognition (CVPR)}, pages 994--1003, 2018.

\bibitem{I2Ireid_nips2018_FDGAN}
Yixiao Ge, Zhuowan Li, Haiyu Zhao, Guojun Yin, Shuai Yi, Xiaogang Wang, et~al.
\newblock Fd-gan: Pose-guided feature distilling gan for robust person
  re-identification.
\newblock In {\em Proceedings of the Advances in Neural Information Processing
  Systems (NeurIPS)}, pages 1222--1233, 2018.

\bibitem{disentangle_nips2018_CDD}
Abel Gonzalez-Garcia, Joost van~de Weijer, and Yoshua Bengio.
\newblock Image-to-image translation for cross-domain disentanglement.
\newblock In {\em Proceedings of the Advances in Neural Information Processing
  Systems (NeurIPS)}, pages 1287--1298, 2018.

\bibitem{GAN}
Ian Goodfellow, Jean Pouget-Abadie, Mehdi Mirza, Bing Xu, David Warde-Farley,
  Sherjil Ozair, Aaron Courville, and Yoshua Bengio.
\newblock Generative adversarial nets.
\newblock In {\em Proceedings of the Advances in Neural Information Processing
  Systems (NeurIPS)}, pages 2672--2680, 2014.

\bibitem{vireid_aaai2019_HSME}
Yi Hao, Nannan Wang, Jie Li, and Xinbo Gao.
\newblock Hsme: Hypersphere manifold embedding for visible thermal person
  re-identification.
\newblock In {\em Proceedings of the AAAI Conference on Artificial Intelligence
  (AAAI)}, volume~33, pages 8385--8392, 2019.

\bibitem{RESNET}
Kaiming He, Xiangyu Zhang, Shaoqing Ren, and Jian Sun.
\newblock Deep residual learning for image recognition.
\newblock In {\em Proceedings of the IEEE Conference on Computer Vision and
  Pattern Recognition (CVPR)}, pages 770--778, 2016.

\bibitem{TRIP}
Alexander Hermans, Lucas Beyer, and Bastian Leibe.
\newblock In defense of the triplet loss for person re-identification.
\newblock {\em arXiv preprint arXiv:1703.07737}, 2017.

\bibitem{disentangle_eccv2018_MUNIT}
Xun Huang, Ming-Yu Liu, Serge Belongie, and Jan Kautz.
\newblock Multimodal unsupervised image-to-image translation.
\newblock In {\em Proceedings of the European Conference on Computer Vision
  (ECCV)}, pages 172--189, 2018.

\bibitem{disentangle_cvpr2019_styleGAN}
Tero Karras, Samuli Laine, and Timo Aila.
\newblock A style-based generator architecture for generative adversarial
  networks.
\newblock In {\em Proceedings of the IEEE Conference on Computer Vision and
  Pattern Recognition (CVPR)}, pages 4401--4410, 2019.

\bibitem{ADAM}
Diederik~P Kingma and Jimmy Ba.
\newblock Adam: A method for stochastic optimization.
\newblock {\em arXiv preprint arXiv:1412.6980}, 2014.

\bibitem{LOMO}
Shengcai Liao, Yang Hu, Xiangyu Zhu, and Stan~Z Li.
\newblock Person re-identification by local maximal occurrence representation
  and metric learning.
\newblock In {\em Proceedings of the IEEE Conference on Computer Vision and
  Pattern Recognition (CVPR)}, pages 2197--2206, 2015.

\bibitem{MLBP}
Shengcai Liao and Stan~Z Li.
\newblock Efficient psd constrained asymmetric metric learning for person
  re-identification.
\newblock In {\em Proceedings of the IEEE International Conference on Computer
  Vision (ICCV)}, pages 3685--3693, 2015.

\bibitem{GSM}
Liang Lin, Guangrun Wang, Wangmeng Zuo, Xiangchu Feng, and Lei Zhang.
\newblock Cross-domain visual matching via generalized similarity measure and
  feature learning.
\newblock {\em IEEE Transactions on Pattern Analysis and Machine Intelligence
  (TPAMI)}, 39(6):1089--1102, 2016.

\bibitem{I2Ireid_cvpr2018_trans}
Jinxian Liu, Bingbing Ni, Yichao Yan, Peng Zhou, Shuo Cheng, and Jianguo Hu.
\newblock Pose transferrable person re-identification.
\newblock In {\em Proceedings of the IEEE Conference on Computer Vision and
  Pattern Recognition (CVPR)}, pages 4099--4108, 2018.

\bibitem{I2Ireid_cvpr2019_ATNET}
Jiawei Liu, Zheng-Jun Zha, Di Chen, Richang Hong, and Meng Wang.
\newblock Adaptive transfer network for cross-domain person re-identification.
\newblock In {\em Proceedings of the IEEE Conference on Computer Vision and
  Pattern Recognition (CVPR)}, pages 7202--7211, 2019.

\bibitem{disentangle_cvpr2018_reid}
Liqian Ma, Qianru Sun, Stamatios Georgoulis, Luc Van~Gool, Bernt Schiele, and
  Mario Fritz.
\newblock Disentangled person image generation.
\newblock In {\em Proceedings of the IEEE Conference on Computer Vision and
  Pattern Recognition (CVPR)}, pages 99--108, 2018.

\bibitem{dataset_RegDB}
Dat Nguyen, Hyung Hong, Ki Kim, and Kang Park.
\newblock Person recognition system based on a combination of body images from
  visible light and thermal cameras.
\newblock {\em Sensors}, 17(3):605, 2017.

\bibitem{I2Ireid_eccv2018_PNGAN}
Xuelin Qian, Yanwei Fu, Tao Xiang, Wenxuan Wang, Jie Qiu, Yang Wu, Yu-Gang
  Jiang, and Xiangyang Xue.
\newblock Pose-normalized image generation for person re-identification.
\newblock In {\em Proceedings of the European Conference on Computer Vision
  (ECCV)}, pages 650--667, 2018.

\bibitem{rgb-ir}
M~Saquib Sarfraz and Rainer Stiefelhagen.
\newblock Deep perceptual mapping for cross-modal face recognition.
\newblock {\em International Journal of Computer Vision (IJCV)},
  122(3):426--438, 2017.

\bibitem{disentangle_cvpr2019_fineGAN}
Krishna~Kumar Singh, Utkarsh Ojha, and Yong~Jae Lee.
\newblock Finegan: Unsupervised hierarchical disentanglement for fine-grained
  object generation and discovery.
\newblock In {\em Proceedings of the IEEE Conference on Computer Vision and
  Pattern Recognition (CVPR)}, pages 6490--6499, 2019.

\bibitem{reid_cvpr2019_DIMN}
Jifei Song, Yongxin Yang, Yi-Zhe Song, Tao Xiang, and Timothy~M Hospedales.
\newblock Generalizable person re-identification by domain-invariant mapping
  network.
\newblock In {\em Proceedings of the IEEE Conference on Computer Vision and
  Pattern Recognition (CVPR)}, pages 719--728, 2019.

\bibitem{reid_cvpr2019_SPT}
Sijie Song, Wei Zhang, Jiaying Liu, and Tao Mei.
\newblock Unsupervised person image generation with semantic parsing
  transformation.
\newblock In {\em Proceedings of the IEEE Conference on Computer Vision and
  Pattern Recognition (CVPR)}, pages 2357--2366, 2019.

\bibitem{SVDNET}
Yifan Sun, Liang Zheng, Weijian Deng, and Shengjin Wang.
\newblock Svdnet for pedestrian retrieval.
\newblock In {\em Proceedings of the IEEE International Conference on Computer
  Vision (ICCV)}, pages 3800--3808, 2017.

\bibitem{PCB}
Yifan Sun, Liang Zheng, Yi Yang, Qi Tian, and Shengjin Wang.
\newblock Beyond part models: Person retrieval with refined part pooling (and a
  strong convolutional baseline).
\newblock In {\em Proceedings of the European Conference on Computer Vision
  (ECCV)}, pages 480--496, 2018.

\bibitem{disentangle_cvpr2017_DRGAN}
Luan Tran, Xi Yin, and Xiaoming Liu.
\newblock Disentangled representation learning gan for pose-invariant face
  recognition.
\newblock In {\em Proceedings of the IEEE Conference on Computer Vision and
  Pattern Recognition (CVPR)}, pages 1415--1424, 2017.

\bibitem{disentangle_cvpr2019_face_age}
Hao Wang, Dihong Gong, Zhifeng Li, and Wei Liu.
\newblock Decorrelated adversarial learning for age-invariant face recognition.
\newblock In {\em Proceedings of the IEEE Conference on Computer Vision and
  Pattern Recognition (CVPR)}, pages 3527--3536, 2019.

\bibitem{vireid_cvpr2019_D2RL}
Zhixiang Wang, Zheng Wang, Yinqiang Zheng, Yung-Yu Chuang, and Shin'ichi Satoh.
\newblock Learning to reduce dual-level discrepancy for infrared-visible person
  re-identification.
\newblock In {\em Proceedings of the IEEE Conference on Computer Vision and
  Pattern Recognition (CVPR)}, pages 618--626, 2019.

\bibitem{I2Ireid_cvpr2018_PTGAN}
Longhui Wei, Shiliang Zhang, Wen Gao, and Qi Tian.
\newblock Person transfer gan to bridge domain gap for person
  re-identification.
\newblock In {\em Proceedings of the IEEE Conference on Computer Vision and
  Pattern Recognition (CVPR)}, pages 79--88, 2018.

\bibitem{vireid_iccv2017_SYSU}
Ancong Wu, Wei-Shi Zheng, Hong-Xing Yu, Shaogang Gong, and Jianhuang Lai.
\newblock Rgb-infrared cross-modality person re-identification.
\newblock In {\em Proceedings of the IEEE International Conference on Computer
  Vision (ICCV)}, pages 5380--5389, 2017.

\bibitem{sensor}
Xuezhi Xiang, Ning Lv, Zeting Yu, Mingliang Zhai, and Abdulmotaleb El~Saddik.
\newblock Cross-modality person re-identification based on dual-path
  multi-branch network.
\newblock {\em IEEE Sensors Journal}, 2019.

\bibitem{reid_cvpr2019_PAUL}
Qize Yang, Hong-Xing Yu, Ancong Wu, and Wei-Shi Zheng.
\newblock Patch-based discriminative feature learning for unsupervised person
  re-identification.
\newblock In {\em Proceedings of the IEEE Conference on Computer Vision and
  Pattern Recognition (CVPR)}, pages 3633--3642, 2019.

\bibitem{reid_cvpr2019_RAM}
Wenjie Yang, Houjing Huang, Zhang Zhang, Xiaotang Chen, Kaiqi Huang, and Shu
  Zhang.
\newblock Towards rich feature discovery with class activation maps
  augmentation for person re-identification.
\newblock In {\em Proceedings of the IEEE Conference on Computer Vision and
  Pattern Recognition (CVPR)}, pages 1389--1398, 2019.

\bibitem{vireid_aaai2018_TONE}
Mang Ye, Xiangyuan Lan, Jiawei Li, and Pong~C Yuen.
\newblock Hierarchical discriminative learning for visible thermal person
  re-identification.
\newblock In {\em Proceedings of the AAAI Conference on Artificial Intelligence
  (AAAI)}, 2018.

\bibitem{vireid_tifs2019_eBDTR}
Mang Ye, Xiangyuan Lan, Zheng Wang, and Pong~C Yuen.
\newblock Bi-directional center-constrained top-ranking for visible thermal
  person re-identification.
\newblock {\em IEEE Transactions on Information Forensics and Security}, 2019.

\bibitem{vireid_ijcai2018_BCTR}
Mang Ye, Zheng Wang, Xiangyuan Lan, and Pong~C Yuen.
\newblock Visible thermal person re-identification via dual-constrained
  top-ranking.
\newblock In {\em Proceedings of the International Joint Conference on
  Artificial Intelligence (IJCAI)}, pages 1092--1099, 2018.

\bibitem{disentangle_arxiv2019_illumination}
Zelong Zeng, Zhixiang Wang, Zheng Wang, Yung-Yu Chuang, and Shin'ichi Satoh.
\newblock Illumination-adaptive person re-identification.
\newblock {\em arXiv preprint arXiv:1905.04525}, 2019.

\bibitem{disentangle_cvpr2019_gait}
Ziyuan Zhang, Luan Tran, Xi Yin, Yousef Atoum, Xiaoming Liu, Jian Wan, and
  Nanxin Wang.
\newblock Gait recognition via disentangled representation learning.
\newblock In {\em Proceedings of the IEEE Conference on Computer Vision and
  Pattern Recognition (CVPR)}, pages 4710--4719, 2019.

\bibitem{reid_cvpr2019_CASN}
Meng Zheng, Srikrishna Karanam, Ziyan Wu, and Richard~J Radke.
\newblock Re-identification with consistent attentive siamese networks.
\newblock In {\em Proceedings of the IEEE Conference on Computer Vision and
  Pattern Recognition (CVPR)}, pages 5735--5744, 2019.

\bibitem{disentangle_cvpr2019_DGNET}
Zhedong Zheng, Xiaodong Yang, Zhiding Yu, Liang Zheng, Yi Yang, and Jan Kautz.
\newblock Joint discriminative and generative learning for person
  re-identification.
\newblock In {\em Proceedings of the IEEE Conference on Computer Vision and
  Pattern Recognition (CVPR)}, pages 2138--2147, 2019.

\bibitem{cycleGAN}
Jun-Yan Zhu, Taesung Park, Phillip Isola, and Alexei~A Efros.
\newblock Unpaired image-to-image translation using cycle-consistent
  adversarial networks.
\newblock In {\em Proceedings of the IEEE International Conference on Computer
  Vision (ICCV)}, pages 2223--2232, 2017.

\bibitem{reid_cvpr2019_PATB}
Zhen Zhu, Tengteng Huang, Baoguang Shi, Miao Yu, Bofei Wang, and Xiang Bai.
\newblock Progressive pose attention transfer for person image generation.
\newblock In {\em Proceedings of the IEEE Conference on Computer Vision and
  Pattern Recognition (CVPR)}, pages 2347--2356, 2019.

\end{thebibliography}
}

\end{document}